\documentclass[fleqn]{sig-alternate-05-2015}

\usepackage[
  pass,
]{geometry}

\usepackage{balance}
\usepackage{bbm}
\usepackage{hyperref}
\usepackage{caption}
\usepackage{color}
\usepackage{xspace}
\usepackage[utf8x]{inputenc}
\usepackage{graphicx}
\usepackage[lined, ruled, boxed, commentsnumbered]{algorithm2e}
\usepackage{amsmath,amssymb}
\usepackage{booktabs}
\usepackage{verbatim}
\DeclareCaptionType{copyrightbox}
\usepackage{subcaption}
\usepackage{booktabs}
\usepackage{times}
\usepackage{microtype}

\begin{document}

\numberofauthors{1} 
%
\newcommand{\authspace}{\hspace*{.2in}}
\author{
Subhabrata Mukherjee$^\dag$ \authspace  Stephan G\"unnemann$^\ddag$  \authspace Gerhard Weikum$^\dag$\\[2mm]
 \affaddr{$^\dag$Max Planck Institute for Informatics, $^\ddag$Technical University of Munich} \\
 \email{smukherjee@mpi-inf.mpg.de, guennemann@in.tum.de, weikum@mpi-inf.mpg.de}
 }

\title{Continuous Experience-aware Language Model}

\newcommand{\squishlist}{
   \begin{list}{$\bullet$}
    { \setlength{\itemsep}{0pt}      \setlength{\parsep}{3pt}
      \setlength{\topsep}{3pt}       \setlength{\partopsep}{0pt}
      \setlength{\leftmargin}{1.5em} \setlength{\labelwidth}{1em}
      \setlength{\labelsep}{0.5em} } }
\newcommand{\squishlisttwo}{
   \begin{list}{$\bullet$}
    { \setlength{\itemsep}{0pt}    \setlength{\parsep}{0pt}
      \setlength{\topsep}{0pt}     \setlength{\partopsep}{0pt}
      \setlength{\leftmargin}{0.5em} \setlength{\labelwidth}{0.5em}
      \setlength{\labelsep}{0.5em} } }

\newcommand{\squishend}{
    \end{list} 
}

\newcommand{\todo}[1]{
\textcolor{red}{\textbf{(TODO: #1)}}
}

\CopyrightYear{2016} 

\setcopyright{acmcopyright}

\conferenceinfo{KDD '16,}{August 13-17, 2016, San Francisco, CA, USA}

\isbn{978-1-4503-4232-2/16/08}\acmPrice{\$15.00}

\doi{http://dx.doi.org/10.1145/2939672.2939780}

\maketitle

\section*{Abstract}

%
Online review communities are dynamic as users join and leave, adopt new vocabulary, and adapt to evolving trends. 
Recent work has shown that recommender systems benefit from
explicit consideration of user experience. 
However, prior work assumes a fixed number of
discrete experience levels, whereas in reality 
users gain experience and mature continuously over time.

This paper presents a new model that captures
the {\em continuous evolution} of user experience,
and the resulting {\em language model} in reviews and
other posts.
Our model is unsupervised 
and combines principles of 
{Geometric Brownian Motion}, Brownian Motion, 
and Latent Dirichlet Allocation to trace a smooth temporal progression of user experience and
language model respectively.
%
We develop practical algorithms for estimating the
model parameters from data and for inference with
our model (e.g., to recommend items).
Extensive experiments with five real-world datasets show
that our model not only fits data better than
discrete-model baselines, but also outperforms state-of-the-art 
methods for predicting item ratings. 

\vspace*{-1mm}


\printccsdesc

\vspace*{-1.5mm}
\vspace*{-0.5mm}

\section{Introduction}

{\bf Motivation:}
Review communities about
items like movies, cameras, restaurants, beer, newspapers and more
are a key asset for recommender systems. 
State-of-the-art methods harness different signals for predictions: 
 user-user and item-item similarities in addition to user-item ratings. These are 
typically cast into latent factor models \cite{koren2011advances} that exploit 
user-user interactions, user bias, bursty posting behavior and 
other community-level features 
(e.g., \cite{KorenKDD2010,MaWSDM2011,Gunnemann2014}).
%
None of the above methods, however, consider the role of {\em user experience} and the
{\em evolution} of how users mature over time.
These dimensions have been recognized, investigated and incorporated
into recommender models only recently by \cite{mcauleyWWW2013,Subho:ICDM2015}.

{\bf Example:}
Experienced users often appreciate certain facets of an item differently from novices
and amateurs. 
As users gain experience they mature, and may later appreciate these more
intricate facets. 
Consider the following reviews about Christopher Nolan movies. The
facet of interest is the (non-linear) narrative style. 
\squishlist
\item {\small \em User 1 on Memento (2001): ``A story does not become interesting if told backwards.''}
\item {\small \em User 2 on The Dark Knight (2008): ``Memento was very complicated. The Dark Knight was flawless. Heath Ledger rocks!''}
\item {\small \em User 3 on Inception (2010): ``Inception is a triumph of style over substance. It is complex only in a structural way, not in terms of plot. It doesn't unravel in the way Memento does.''}
\squishend
\noindent The first user does not appreciate complex narratives. The second user prefers simpler
blockbusters. The third user seems to appreciate the non-linear narration style
of Inception and, more so of Memento.
In terms of maturity, we would consider User 3 to be more experienced in the underlying facet,
and use this assessment when generating future recommendations to her or
similar users.

{\bf State-of-the-Art and its Limitations:}
The evolution of user experience and how it affects ratings
has first been studied in~\cite{mcauleyWWW2013,Subho:ICDM2015}. 
However, these works make the simplifying assumption that user experience is
categorical with discrete levels (e.g. $[1, 2, 3, \ldots, E]$), and that
users progress from one level to the next in a discrete manner. 
As an artifact of this assumption, the experience level of a user changes abruptly by one transition. 
Also, an undesirable consequence of the discrete model is that
all users at the same level of experience are treated similarly, although their maturity could still be far apart (if
we had a continuous scale of measuring experience). Therefore, the assumption of {\em exchangeability} of reviews, for the latent factor model in \cite{Subho:ICDM2015}, for users at the same level of experience may not hold as the language model changes.

The prior work~\cite{mcauleyWWW2013} assumes user {\em activity} (e.g., number of reviews) to play a major role in experience evolution, which biases the model towards highly active users
(as opposed to an experienced person who posts only once in a while).
In contrast, our own prior work \cite{Subho:ICDM2015} captures {\em interpretable} evidence for a user's experience level using her vocabulary, cast into a 
language model with latent facets.
However, this approach also exhibits the drawbacks of discrete levels of
experience, as discussed above. 

The current paper overcomes these limitations by modeling
the evolution of user experience, and the corresponding 
language model, as a {\em continuous-time} stochastic process. We model time {\em explicitly} in this work, in contrast to the prior works. 

%


Other prior work on item recommendation that considered review texts (e.g., \cite{mcauleyRecSys2013},~\cite{wang2011},~\cite{mukherjeeSDM2014})
did this with the sole perspective of learning topical similarities in a
static snapshot-oriented manner, without considering time at all.





{\bf Approach and Technical Challenges:}
This paper is the first work to develop a continuous-time model of user experience 
and language evolution.
Unlike prior work, we do not rely on explicit features like ratings or number of reviews.  
Instead, 
we capture a user's experience by a latent language model learned from
the user-specific vocabulary in her review texts.
We present a generative model where the user's experience and language model evolve according to a Geometric Brownian Motion (GBM) and Brownian Motion process, respectively. 
Analysis of the GBM trajectory of users offer interesting insights;
for instance, users who reach a high level of experience progress faster 
than those who do not, and also exhibit a comparatively higher variance.
Also, the number of reviews written by a user does not have a strong influence, 
unless they are written over a long period of time.

The facets in our model (e.g., narrative style, actor performance, etc. for  movies) are generated using Latent Dirichlet Allocation. 
User experience and item facets are latent variables,
whereas the observables are
 {\em words}  at explicit {\em timepoints} 
in user reviews. 

The parameter estimation and inference for our model are challenging since 
we combine discrete multinomial distributions (generating words per review) with a continuous Brownian Motion process for the language models' evolution, and a continuous Geometric Brownian Motion (GBM) process for the user experience. 

{\bf Contributions:}
To solve this technical challenge, we present an inference method consisting of three steps: a) estimation of user experience from a user-specific GBM using the 
Metropolis Hastings algorithm, 
b) estimation of the language model evolution by Kalman Filter, and c) estimation of latent facets using Gibbs sampling. Our experiments, with
real-life data from five different communities on movies, food, beer and
news media, show that the three components {\em coherently} work together and
yield a better fit of the data (in terms of log-likelihood) than the previously
best models with discrete experience levels.
%
We also achieve an improvement of ca. $11\%$ to $36\%$ for the mean squared
error for predicting user-specific ratings of items compared to the baseline of \cite{mcauleyWWW2013,Subho:ICDM2015}.
Finally, we present a use-case study with a news-media community, where experience-aware models can be used to identify experienced {\em citizen journalists} --- where our method performs well in capturing user maturity.

In summary, our contributions are:
\squishlist
	\item Model: We devise a probabilistic model for tracing {\em continuous} evolution of  {\em user experience}, combined with a language model for facets that explicitly captures smooth evolution over time.
	\item Algorithm: We introduce an effective learning algorithm, that infers each users' experience progression, time-sensitive language models, 
and latent facets of each word. 
	\item Experiments: We perform extensive experiments with five real-word
 datasets, together comprising of $12.7$ million ratings from $0.9$ million users on $0.5$ million items, and demonstrate substantial improvements of our method over state-of-the-art baselines.
\squishend 

The rest of the paper is organized as follows.
Section 2 introduces the fundamental components for modeling the continuous evolution of user experience and language.
Section 3 presents the generative model for the joint evolution of
facets, experience, and language, and also our inference methods based
on Markov Chain Monte Carlo (MCMC) sampling.
Section 4 shows our experimental results, comparing our model against a variety of baselines.
Section 5 discusses the use-case study, followed by related prior work.

\section{Model Components}

 \begin{figure*}
\begin{subfigure}{.49\textwidth}
  \centering
  \includegraphics[width=\linewidth]{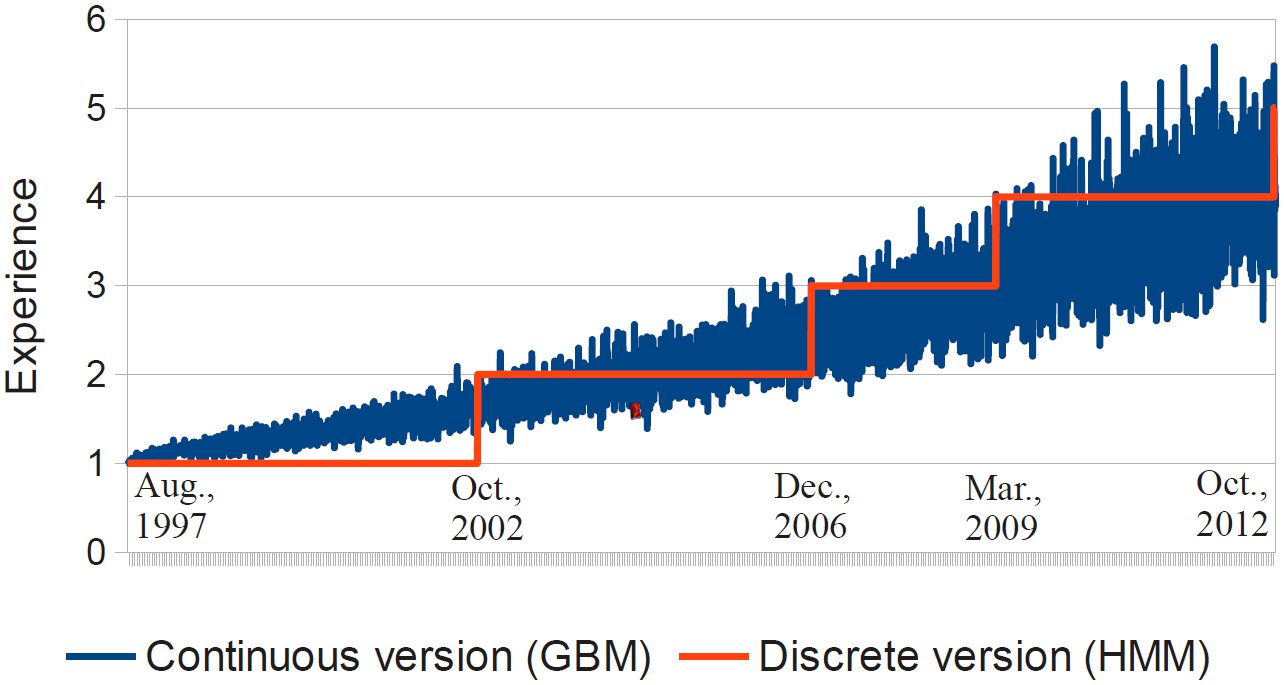}
  \caption{Evolution of an experienced user.}
  \label{fig:exp1}
\end{subfigure}%
\begin{subfigure}{.49\textwidth}
  \centering
  \includegraphics[width=1.05\linewidth]{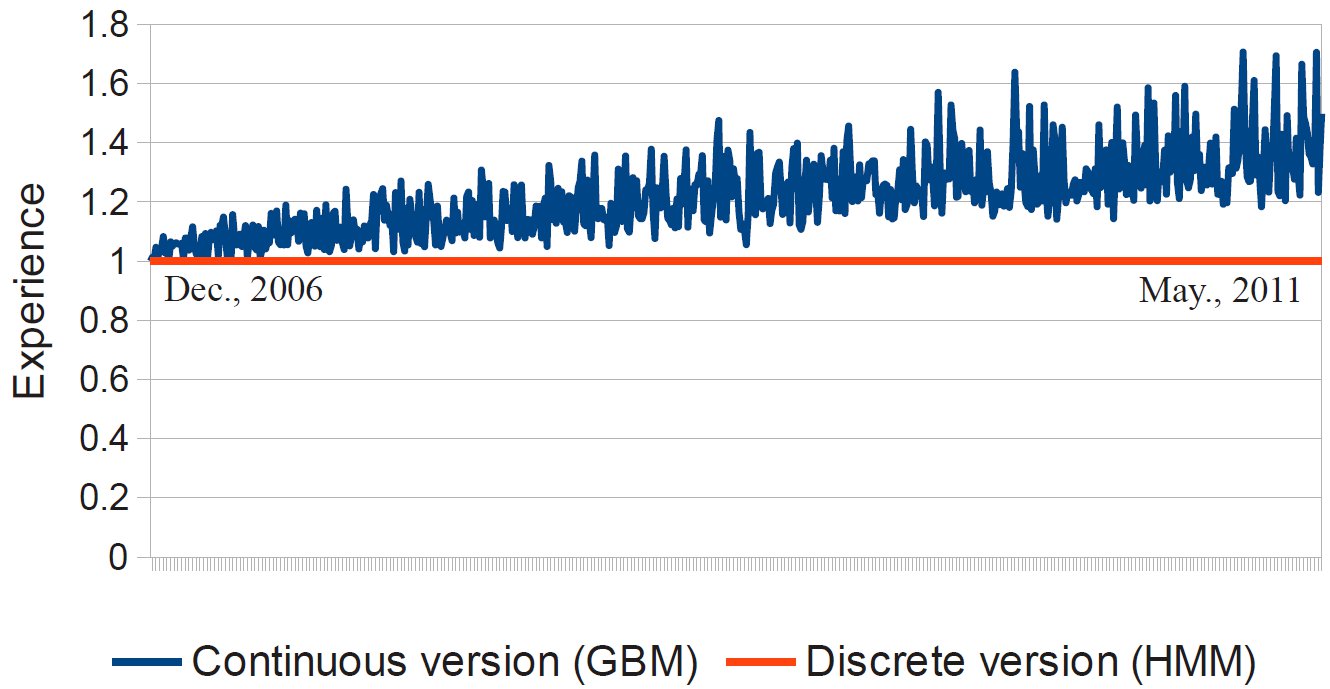}
  \caption{Evolution of an amateur user.}
  \label{fig:exp2}
\end{subfigure}
\caption{Discrete state and continuous state experience evolution of some typical users from the BeerAdvocate community.}
\label{fig:exp}
\end{figure*}

 \vspace*{-2mm}\subsection{Importance of Time}\label{sec:time}
Previous works on experience evolution~\cite{mcauleyWWW2013,Subho:ICDM2015} 
model time only implicitly by assuming the (discrete) latent experience to progress from one review to the next.
%
%
%
In contrast, this current paper models time explicitly, and allows experience to {\em continuously} evolve over time --- so that we are able to trace the joint evolution of experience, and vocabulary. This is challenging as the discrete Multinomial distribution based language model (to generate words) needs to be combined with a continuous stochastic process for experience evolution.

We use {two} levels of temporal granularity. Since experience is naturally {continuous}, it is beneficial to model its evolution at a very fine resolution (say, minutes or hours). 
On the other hand, the language model has a much coarser granularity (say, days, weeks or months). 
We show in Section~\ref{sec:inference} how to smoothly merge the two granularities using continuous-time models. 
Our model for language evolution is motivated by the seminal work of Wang and Blei et al.~\cite{BleiCTM},
with major differences and extensions. 
%
In the following subsections, we formally introduce the two components affected by time: the experience evolution and the language model evolution.


\subsection{Continuous Experience Evolution}
 
Prior works~\cite{mcauleyWWW2013,Subho:ICDM2015} model experience as a discrete random variable . At each timepoint, a user is allowed to stay at level $l$, or move to level $l+1$. As a result the transition is abrupt when the user switches levels. Also, the model does not distinguish between users at the same level of experience, (or even for the same user at beginning or end of a level) even though their experience can be quite far apart (if measured in a continuous scale). For instance, in Figure~\ref{fig:exp2} the language model uses the same set of parameters as long as the user stays at level $1$, although the language model changes. In order to address these issues, our goal is to develop a continuous experience evolution model with the following requirements:
\begin{itemize}
 \item The experience value is always positive.
 \item Markovian assumption for the continuous-time process: The experience value at any time $t$ depends only on the value at the {\em most recent 
observed 
time prior to $t$}.
 \item Drift: It has an overall {\em trend} to increase over time.
 \item Volatility:  The evolution may not be smooth with occasional volatility. For instance, an experienced user may write a series of expert reviews, followed by a sloppy one.
\end{itemize}

To capture all of these aspects, we model each user's experience as a {\em Geometric Brownian Motion} (GBM) process (also known as Exponential Brownian Motion). 

GBM is a natural continuous state alternative to the discrete-state space based Hidden Markov Model (HMM) used in our previous work~\cite{Subho:ICDM2015}. Figure~\ref{fig:exp} shows a real-world example of the evolution of an experienced and amateur user in the BeerAdvocate community, as traced by our proposed model --- along with that of its discrete counterpart from our previous work. The GBM is a stochastic process used to model population growth, financial processes like stock price behavior (e.g., Black-Scholes model) with random noise. It is a continuous time stochastic process, where the logarithm of the random variable (say, $X_t$) follows Brownian Motion with a {\em volatility} and {\em drift}.
Formally, a stochastic process $X_t$, with an arbitrary initial value $X_0$, for $t \in [0,\infty)$ is said to follow Geometric Brownian Motion, if it satisfies the following Stochastic Differential Equation (SDE)~\cite{gbm}:
\begin{equation}
\label{eq:sde}
 dX_t = \mu X_t dt + \sigma X_t dW_t
\end{equation}
where, $W_t$ is a Wiener process (Standard Brownian Motion); $\mu \in \mathbb{R}$ and $\sigma \in (0, \infty)$ are constants called the {\em percentage trend} and {\em percentage volatility} respectively. The former captures deterministic trends, whereas the latter captures unpredictable events occurring during the motion.

In a Brownian Motion trajectory, $\mu X_tdt$ and $\sigma X_t dW_t$ capture the ``trend'' and ``volatility'', as is required for experience evolution. However, in real life communities  each user might show a different experience evolution; therefore our model considers a {\em multivariate} version of this GBM -- we model one trajectory {\em per-user}. Correspondingly, during the inference process we learn $\mu_u$ and $\sigma_u$ for each user $u$.

\emph{Properties:} A straightforward application of {\em It\^{o}'s} formula yields the following analytic solution to the above SDE (Equation~\ref{eq:sde}):
\begin{equation}
\begin{aligned}
 X_t &= X_0 \ exp\big((\mu - \frac{\sigma^2}{2})t + \sigma W_t\big)\\
 \end{aligned}
\end{equation}

Since $log(X_t)$ follows a Normal distribution, $X_t$ is Log-Normally distributed with mean $\big( log(X_0) + (\mu - \frac{\sigma^2}{2})t \big)$ and variance $\sigma \sqrt{t}$. The probability density function $f_t(x)$, for $x \in (0, \infty)$, is given by:
\begin{equation}
\label{eq:ln}
 f_t(x) = \frac{1}{\sqrt{2 \pi t}\sigma x} exp \bigg(-\frac{\big(log(x) - log\ (x_0) - (\mu - \frac{\sigma^2}{2})t\big)^2}{2\sigma^2t}\bigg)
\end{equation}

It is easy to show that GBM has the Markov property. Consider $U_t = (\mu - \frac{\sigma^2}{2})t + \sigma W_t$.
\begin{equation}
\begin{aligned}
 X_{t+h} &= X_0 exp(U_{t+h})\\
&= X_0 exp (U_t + U_{t+h} - U_t)\\
&= X_0 exp (U_t)exp(U_{t+h} - U_t)\\
&= X_t exp (U_{t+h} - U_t)
\end{aligned}
\end{equation}

Therefore, future states depend only on the future increment of the Brownian Motion, which satisfies our requirement for experience evolution. Also, for $X_0 > 0$, the GBM process is always positive. {\em Note} that the start time of the GBM of {\em each} user is relative to her first review in the community.

\subsection{Experience-aware Language 
Evolution}

Once the experience values for each user are generated from a Log-Normal distribution (more precisely: the experience of the user at the times when she wrote each review), we develop the language model whose parameters evolve according to the Markov property for experience evolution.

As users get more experienced, they use more sophisticated words to express a concept. For instance, 
experienced cineastes refer to a movie's ``protagonist'' whereas amateur movie lovers talk about the ``hero''.
Similarly, in a Beer review community (e.g., BeerAdvocate, RateBeer) experts use more {\em fruity} words to describe a beer like ``caramel finish, coffee roasted vanilla'', ``and citrus hops''. Facet preferences of users also evolve with experience. For example, users at a high level of experience prefer ``hoppiest'' beers which are considered too ``bitter'' by amateurs~\cite{mcauleyWWW2013}. 
Encoding explicit time in our model allows us to trace the evolution of vocabulary and trends {\em jointly} on the temporal and experience dimension.

{\noindent \bf Latent Dirichlet Allocation (LDA):} In the traditional LDA process \cite{Blei2003LDA}, a document is assumed to have a distribution over $Z$ facets (a.k.a. topics) $\beta_{1:Z}$, and each of the facets has a distribution over words from a fixed vocabulary collection. The per-facet word (a.k.a topic-word) distribution $\beta_{z}$ is drawn from a Dirichlet distribution, and words $w$ are generated from a Multinomial($\beta_z$).

The process assumes that documents are drawn {\em exchangeably} from the same set of facets. However, this process neither takes experience nor the evolution of the facets over {\em time} into account. 

{\noindent \bf Discrete Experience-aware LDA:} Our previous work~\cite{Subho:ICDM2015} incorporates a layer for \emph{experience} in the above process. The user experience is manifested in the set of facets that the user chooses to write on, and the vocabulary and writing style used in the reviews. The experience levels were drawn from a \emph{Hidden Markov Model} (HMM). 
The reviews were assumed to be exchangeable for a user at the same level of experience -- an assumption which generally may not hold; since the language model of a user at the same discrete experience level may be different at different points in time (refer to Figure~\ref{fig:exp2}) (if we had a continuous scale for measuring experience). The process considers {\em time} only {\em implicitly} via the transition of the latent variable for experience.

{\noindent \bf Continuous Time LDA:} The seminal work of Blei et. al in~\cite{BleiDTM,BleiCTM} captures evolving content, for instance, in scholarly journals and news articles where the themes evolve over time, by considering time {\em explicitly} in the generative LDA process. Our language model evolution is motivated by their Continuous Time Dynamic Topic Model~\cite{BleiDTM}, with the major difference that the facets, in our case, evolve over both {\em time} and {\em experience}. 

{\noindent \bf Continuous Experience-aware LDA (this work):} Since the assumption of exchangeability of documents at the same level of experience of a user may not hold, we want the language model to explicitly evolve over experience and time.
To incorporate the effect of changing experience levels, our goal is to condition the parameter evolution of $\beta$ on the experience progression.
%
%
%
%

In more detail, for the language model evolution, we desire the following properties: 
\begin{itemize}
 \item It should {\em smoothly} evolve over time preserving the Markov property of experience evolution.
 \item Its variance should {\em linearly increase} with the  {\em experience change} between successive timepoints. This entails that if the experience of  a user does not change between successive timepoints, the language model remains almost the same. 
\end{itemize}


To incorporate the temporal aspects of data, in our model, we use multiple distributions $\beta_{t,z}$ for each time $t$ and facet $z$. Furthermore, to capture the smooth temporal evolution of the facet language model, we need to chain the different distributions to sequentially evolve over time $t$: the distribution $\beta_{t,z}$ should affect the distribution $\beta_{t+1,z}$.

Since the traditional parametrization of a Multinomial distribution via its mean parameters is not amenable to sequential modeling, and inconvenient to work with in gradient based optimization -- since any gradient step requires the projection to the feasible set, the simplex --- we follow a similar approach as \cite{BleiCTM}: instead of operating on the mean parameters, we consider the natural parameters of the Multinomial. The natural parameters are unconstrained and, thus, enable an easier sequential modeling.

From now on, we denote with $\beta_{t,z}$ the natural parameters of the Multinomial at time $t$ for facet $z$. For {\it identifiability} one of the parameters $\beta_{t,z,w}$ needs to be fixed at zero. By applying the following mapping we can obtain back the mean parameters that are located on the simplex:
\begin{equation}
\label{eq:pi}
\pi(\beta_{t,z,w}) = \frac{exp(\beta_{t,z,w})}{1 + \sum_{w=1}^{V-1} exp(\beta_{t,z,w})}
\end{equation}

Using the natural parameters, we can now define the facet-model evolution:
The underlying idea is that strong changes in the users' experience can lead to strong changes in the language model, while low changes should lead to only few changes. To capture this effect, let $l_{t,w}$ denote the average experience of a word $w$ at time $t$ (e.g. the value of $l_{t,w}$ is high if many experienced users have used the word). That is, $l_{t,w}$ is given by the average experience of all the reviews $D_t$ containing the word $w$ at time $t$.
 
 \begin{equation}
 \label{eq:word-exp}
 l_{t,w} =  \frac{\sum_{d\in D_t: w \in d} e_d}{|D_t|}
 \end{equation}
 where, $e_d$ is the experience value of review $d$ (i.e. the experience of user $u_d$ at the time of writing the review).
 

The language model evolution is then modeled as:
 \begin{equation}
 \label{eq:beta}
 \beta_{t,z,w} \sim Normal (\beta_{t-1,z,w}, \sigma \cdot |l_{t,w} - l_{t-1,w}|) 
 \end{equation}
Here, we simply follow the idea of a standard dynamic system with Gaussian noise, where the mean is the value at the previous timepoint, and the variance increases linearly with increasing change in the experience. Thereby, the desired properties of the language model evolution are ensured.



\section{Joint Model for Experience-\\ Language Evolution}
\label{sec:inference}

\subsection{Generative Process}
Consider a corpus $D=\{d_1,\ldots,d_D\}$ of review documents written by a set of users $U$ at timestamps $T$. For each review $d\in D$, we denote $u_d$ as its user, $t'_d$ as the fine-grained timestamp of the review (e.g. minutes or seconds; used for experience evolution) and with $t_d$ the timestamp of coarser granularity (e.g. yearly or monthly; used for language model evolution). 
The reviews are assumed to be ordered by timestamps, i.e. $t{'}_{d_i}<t{'}_{d_j}$ for $i<j$. We denote with $D_t=\{d\in D \mid t_d=t\}$ all reviews written at timepoint $t$.
 Each review $d \in D$ consists of a sequence of $N_d$ words denoted by $d=\{w_1,\ldots ,w_{N_d}\}$, where each word is drawn from a vocabulary $V$ having unique words indexed by $\{1 \dots V\}$. The number of facets corresponds to $Z$.
 
 Let $e_d \in (0, \infty)$ denote the experience value of review $d$. Since each review $d$ is associated with a unique timestamp $t'_d$ and unique user $u_d$, the experience value of a review refers to the experience of the user at the time of writing it.
 %
%
%
In our model, each user $u$ follows her own Geometric Brownian Motion trajectory -- starting time of which is relative to the first review of the user in the community --  parametrized by the mean $\mu_u$, variance $\sigma_u$, and her {\em starting experience} value $s_{0,u}$. As shown in Equation~\ref{eq:ln}, the analytical form of a GBM translates to a Log-Normal distribution with the given mean and variance. We use this user-dependent distribution to generate an experience value $e_d$ for the review $d$ written by her at timestamp $t'_d$. 

Following standard LDA, the facet proportion $\theta_d$ of the review is drawn from a Dirichlet distribution with concentration parameter $\alpha$, and the facet $z_{d,w}$ of each word $w$ in $d$ is drawn from a Multinomial($\theta_d$).
%

Having generated the experience values, we can now generate the language model and individual words in the review.
Here, the language model $\beta_{t,z,w}$ uses the state-transition Equation~\ref{eq:beta}, and the actual word $w$ is based on its facet $z_{d,w}$ and timepoint $t_d$ according to a Multinomial($\pi(\beta_{t_d,z_{d,w}})$), where the transformation $\pi$ is given by Equation~\ref{eq:pi}.

Note that technically, the distribution $\beta_t$ and word $w$ have to be generated simultaneously: for $\beta_t$ we require the terms $l_{t,w}$, which depend on the experience and the words. Thus, we have a joint distribution $P(\beta_t,w|\ldots)$. Since, however, words are \emph{observed} during inference, this dependence is not crucial, i.e.\ $l_{t,w}$ can be computed once the experience values are known using Equation~\ref{eq:word-exp}. 

We use this observation to simplify the notations and illustrations of Algorithm~\ref{algo:1}, which outlines the generative process, and Figure~\ref{fig:2}, which depicts it visually in plate notation for graphical models.

%

 \begin{figure}[h]
 \centering
 \includegraphics[scale=0.4]{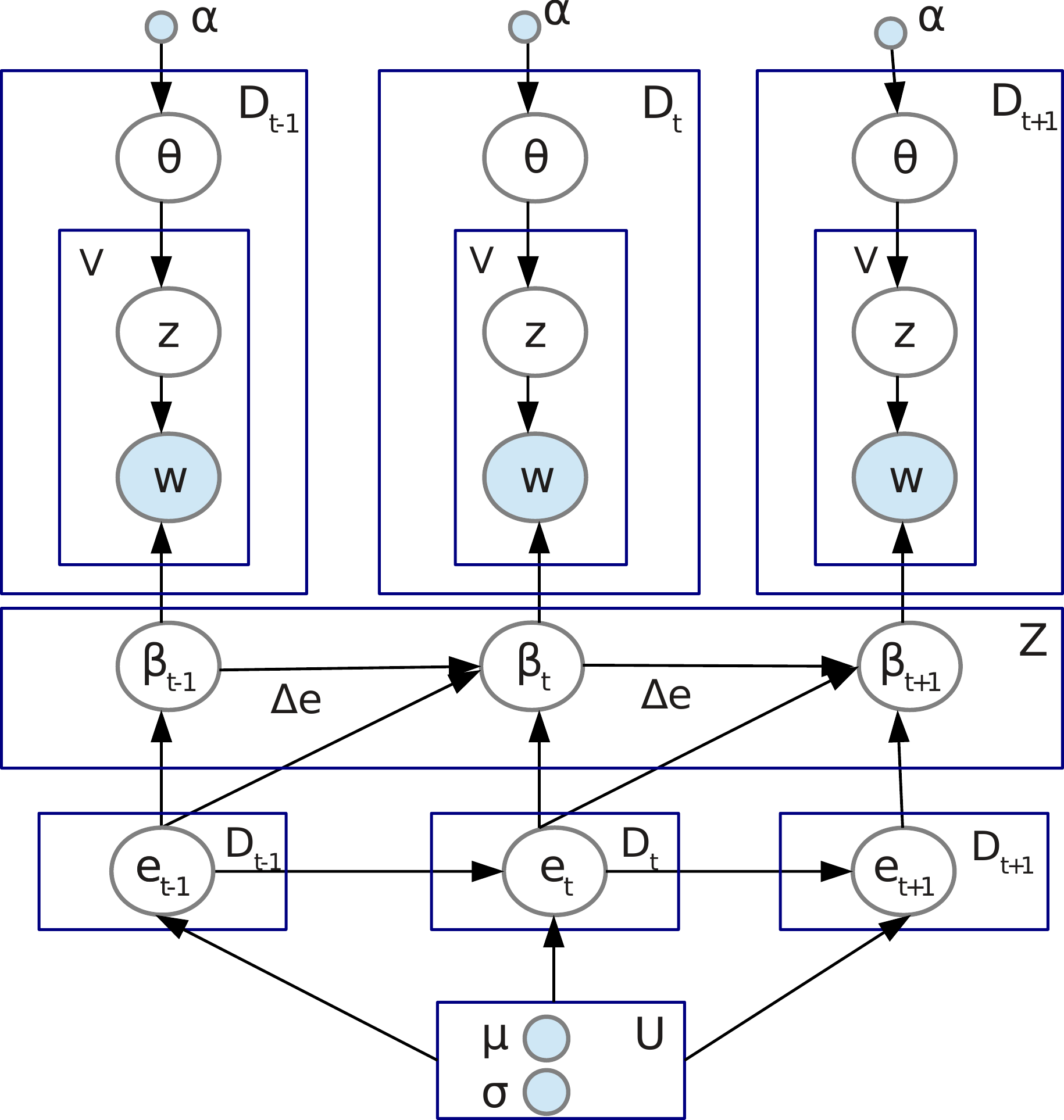}
  \caption{\small Continuous experience-aware language model. Words (shaded in blue), and timestamps (not shown for brevity) are observed.}
  \label{fig:2}
  \vspace{-1em}
 \end{figure}

 \begin{algorithm}[t]
\SetAlgoLined
\DontPrintSemicolon
{\small
1. {\small Set granularity $t$ for language model evolution (e.g., years, months, days)}\;
2. {\small Set granularity for experience evolution, timestamp $t'$ (e.g., minutes, seconds)}\;

\For {each coarse timepoint t} {
\For {each review $d\in D_t$} {
	// retrieve user $u=u_d$ and fine-grained timepoint $t'=t'_d$\;
3. Draw $e_d \sim \text{Log-Normal}((\mu_u-\frac{\sigma_u^2}{2})t{'} +log(s_{0,u}), \sigma_u \sqrt{t{'}})$\;
4. Draw $\theta_d \sim \text{Dirichlet}(\alpha)$\;
\For {each word $w$ in $d$} {
5. Draw $z_{d,w} \sim \text{Multinomial}(\theta_d$)\;
}
}
6. Draw $\beta_{t,z,w} \sim \text{Normal}(\beta_{t-1,z,w}, \sigma \cdot |l_{t,w} - l_{t-1,w}|)$\;
\For {each review $d\in D_t$} {
	\For {each word $w$ in $d$} {
7. Draw $w \sim \text{Multinomial}(\pi(\beta_{t_d,z_{d,w}}))$
}
}
}
	}


\caption{Generative model for continuous experience-aware language model.}
\label{algo:1}
\end{algorithm}

\subsection{Inference}

Let $E, L, Z, T$  and $W$ be the set of experience values of all reviews, experience values of words, facets, timestamps and words in the corpus, respectively. 
In the following, $d$ denotes a review and $j$ indexes a word in it. 
$\theta$ denotes the per-review facet distribution, and $\beta$ the language model respectively. 

The joint probability distribution is given by:

{\small
\setlength{\mathindent}{0cm}
\begin{multline}
 P(E,L, Z,W, \theta, \beta | U, T; \alpha, \langle \mu \rangle, \langle \sigma \rangle) \propto
 \prod_{t \in T} \prod_{d\in D_t}  P(e_d; s_{0,u_d}, \mu_{u_d}, \sigma_{u_d}) \\ 
 \boldsymbol{\cdot} \bigg( P(\theta_d ; \alpha) \cdot \prod_{j=1}^{N_d} P(z_{d,j} | \theta_d)   \cdot P(w_{d,j} | \pi(\beta_{z_{d,j},t})) \bigg) \\
 \boldsymbol{\cdot}  \bigg( \prod_{z \in Z} \prod_{w \in W} P (l_{t,w}; e_d)\cdot P(\beta_{t,z,w}; \beta_{t-1,z,w}, \sigma \cdot |l_{t,w} - l_{t-1,w}|) \bigg) 
\end{multline}
}

The exact computation of the above distribution is intractable, and we have to resort to approximate inference. Exploiting conjugacy of the Multinomial and Dirichlet distributions, we can integrate out $\theta$ from the above distribution. Assuming $\theta$ has been integrated out, we can decompose the joint distribution as:

{
\small
\setlength{\mathindent}{0cm}
\begin{equation}
  P(Z, \beta, E, L | W, T) \propto P(Z, \beta | W, T) \cdot P( E | Z, \beta, W, T) \cdot P(L | E, W, T)
\end{equation}
}

The above decomposition makes certain conditional independence assumptions in line with our generative process.

\subsubsection{Estimating Facets $Z$}

We use Collapsed Gibbs Sampling~\cite{Griffiths02gibbssampling}, as in standard LDA,
to estimate the conditional distribution for each of the latent facets $z_{d,j}$,
which is computed over the current assignment for all other hidden
variables, after integrating out $\theta$. Let $n(d, z)$ denote the count of the topic $z$ appearing in review $d$.
In the following equation,
 $n(d,.)$  indicates the summation of the above counts over all possible $z\in Z$.  
The subscript $-j$ denotes the value
of a variable excluding the data at the $j^{th}$ position.

The posterior distribution
$P(Z| \beta, W, T; \alpha )$ of the latent variable $Z$ is given by:

\begin{equation}
\begin{aligned}
 & P(z_{d,j} = k | z_{d,-j}, \beta, w_{d,j}, t, d; \alpha) \\
 & \propto \frac{n(d, k) + \alpha}{n(d, .) + Z \cdot \alpha} \boldsymbol{\cdot} P(w_n = w_{d,j}| \beta, t, z_n =k, z_{-n}, w_{-n}) & \\
 & = \frac{n(d, k) + \alpha}{n(d, .) + Z \cdot \alpha} \boldsymbol{\cdot} \pi(\beta_{t,k,w_n})\\
\end{aligned} 
\label{eq:gibbs}
\vspace{-2em}
\end{equation}
where, the transformation $\pi$ is given by Equation~\ref{eq:pi}.

\subsubsection{Estimating Language Model $\beta$}
In contrast to $\theta$, the variable $\beta$ cannot be integrated out by the same process, as Normal and Multinomial distributions are not conjugate. Therefore, we refer to  another approximation technique to estimate $\beta$.

In this work, we use {\em Kalman Filter}~\cite{kalman} to model the sequential language model evolution. It is widely used to model linear dynamic systems from a series of observed measurements over time, containing statistical noise, that produces robust estimates of unknown variables over a single measurement. It is a continuous analog to the Hidden Markov Model (HMM), where the state space of the latent variables is continuous (as opposed to the discrete state-space HMM); and the observed and latent variables evolve with Gaussian noise. 

We want to estimate the following state-space transition model:
\begin{equation}
\begin{aligned}
\beta_{t,z,w} | \beta_{t-1,z, w} & \sim N (\beta_{t-1,z, w}, \sigma \cdot |l_{t,w} - l_{t-1, w}|)\\
w_{d,j} | \beta_{t,z, w} & \sim Mult(\pi(\beta_{t,z,w})) \text { where, } z=z_{d,j}, t=t_d.\\
\end{aligned} 
\end{equation}

However, unlike standard Kalman Filter, we do not have any {\em observed} measurement of the variables --- due to the presence of {\em latent} facets $Z$. Therefore, we resort to {\em inferred} measurement from the Gibbs sampling process.  

Let $n(t, z, w)$ denote the number of times a given word $w$ is assigned to a facet $z$ at time $t$ in the corpus. Therefore,
\begin{equation}
 \beta^{inf}_{t,z,w} = \pi^{-1} \bigg( \frac{n(t, z, w) + \gamma}{n(t, z, .) + V \cdot \gamma} \bigg)
\end{equation}
where, we use the inverse transformation of $\pi$ given by Equation~\ref{eq:pi}, and 
$\gamma$ is used for smoothing.

\noindent{\bf Update Equations for Kalman Filter: }
Let $p_t$ and $g_t$ denote the {\em prediction error}, and {\em Kalman Gain} at time $t$ respectively. The variance of the process noise and measurement is given by the difference of the experience value of the word observed at two successive timepoints. Following standard Kalman Filter calculations~\cite{kalman}, the predict equations are given by:
\begin{equation}
\label{eq:kalman-predict}
\begin{aligned}
\widehat{\beta}_{t,z,w} & \sim N(\beta_{{t-1},z,w}, \sigma \cdot |l_{t,w} - l_{t-1, w}|)\\
\widehat{p}_t & = p_{t-1} + \sigma \cdot |l_{t-1,w} - l_{t-2, w}|
\end{aligned} 
\end{equation}
and the update becomes:
\begin{equation}
\label{eq:kalman-update}
\begin{aligned}
g_t & = \frac{\widehat{p}_t}{ \widehat{p}_t + \sigma \cdot |l_{t,w} - l_{t-1, w}|}\\
\beta_{t,z,w} & = \widehat{\beta}_{t,z,w} + g_t \cdot (\beta^{inf}_{t,z,w} - \widehat{\beta}_{t,z,w}) \\
p_t & = (1 - g_t) \cdot \widehat{p}_t
\end{aligned} 
\end{equation}
Thus, the new value for $\beta_{t,z,w}$ is given by Eq. \ref{eq:kalman-update}.

If the experience does not change much between two successive timepoints, i.e. the variance is close to zero, the Kalman Filter just emits the counts as estimated by Gibbs sampling (assuming, $P_0 = 1$). This is then similar to the Dynamic Topic Model~\cite{BleiDTM}. Intuitively, the Kalman Filter is smoothing the estimate of Gibbs sampling taking the experience evolution into account.

\subsubsection{Estimating Experience $E$}

The experience value of a review depends on the user and the language model $\beta$. Although we have the state-transition model of $\beta$, the previous process of estimation using Kalman Filter cannot be applied in this case, as there is no observed or inferred value of $E$. Therefore, we resort to
Metropolis Hastings sampling. 
Instead of sampling the $E$'s from the complex true distribution, we use a proposal distribution for sampling the random variables --- followed by an acceptance or rejection of the newly sampled value.
 That is, at each iteration, the algorithm samples a value of a random variable --- where the current estimate depends only on the previous estimate, thereby, forming a Markov chain. 

 Assume all reviews $\{\cdots d_{i-1}, d_i, d_{i+1} \cdots\}$ from all users are sorted according to their timestamps.
 As discussed in Section~\ref{sec:time}, for computational feasibility,
 we use a coarse granularity for the language model $\beta$. For the inference of $E$, however, we need to operate at the fine temporal resolution of the reviews' timestamps (say, in minutes or seconds).
Note that the process defined in Eq.~\eqref{eq:beta} represents the aggregated language model over multiple fine-grained timestamps. Accordingly, its corresponding fine-grained counterpart is $\beta_{t'_{d_i},z,w} \sim Normal (\beta_{t'_{d_{i-1}},z,w}, \sigma \cdot |e_{d_i} - e_{d_{i-1}}|) $ --- now operating on $t'$ and the review's individual experience values.
Since the language model is given (i.e. previously estimated) during the inference of $E$ , we can now easily refer to this fine-grained definition for the Metropolis Hastings sampling.

 


As the proposal distribution for the experience of review $d_i$ at time $t'_{d_i}$ , we select the corresponding user's GBM ($u=u_d$) and sample a new experience value $\widehat{e_{d_i}}$ for the review:
$$\widehat{e}_{d_i} \sim \text{Log-Normal}((\mu_u-\frac{\sigma_u^2}{2})t'_{d_i} +log(s_{0,u}), \sigma_u \sqrt{t'_{d_i}})$$

 
\noindent The language model $\beta_{t'_{d_i}}$ at time $t'_{d_i}$ depends on the language model $\beta_{t'_{d_{i-1}}}$ at time $t'_{d_{i-1}}$, and experience value difference $|e_{d_i} - e_{d_{i-1}}|$ between the two timepoints. Therefore, a change in the experience value at any timepoint affects the language model at the \emph{current} and {next} timepoint, i.e. $\beta_{t^{'}_{d_{i+1}}}$ is affected by $\beta_{t^{'}_{d_{i}}}$, too.

Thus, the acceptance ratio of the Metropolis Hastings sampling becomes:
\begin{multline}
\label{eq:prop}
Q = \prod_{w,z} \bigg[\frac{N(\beta_{t'_b,z,w}; \beta_{t'_a, z, w}, \sigma \cdot | \widehat{e_b} - e_{a}|)}{N(\beta_{t'_b,z,w}; \beta_{t'_a, z, w}, \sigma \cdot | e_b - e_{a}|)} \\
\boldsymbol{\cdot} \frac{N(\beta_{t'_c,z,w}; \beta_{t'_b, z, w}, \sigma \cdot | e_{c} - \widehat{e_b}|)}{N(\beta_{t'_c,z,w}; \beta_{t'_b, z, w}, \sigma \cdot | e_{c} - e_b|)}\bigg]
\end{multline}
where $a=d_{i-1}$, $b=d_{i}$ and $c=d_{i+1}$.
The numerator accounts for the modified distributions affected by the updated experience value, and the denominator discounts the old ones. Note that since the GBM has been used as the proposal distribution, its factor cancels out in the term $Q$.

Overall, the Metropolis Hastings algorithm iterates over the following steps:
{\setlength{\leftmargini}{5mm}
\begin{enumerate}
 \item Randomly pick a review $d$ at time $t'=t'_d$ by user $u=u_d$ with experience $e_d$
 \item Sample $\widehat{e_d} \sim \text{Log-Normal}\bigg((\mu_u-\frac{\sigma_u^2}{2})t' +log(s_{0,u}), \sigma_u \sqrt{t'}\bigg)$
 \item Accept $\widehat{e_d}$ as the new experience with probability $P$\,$=$\,$min(1, Q)$
\end{enumerate}}

\subsubsection{Estimating Parameters for the Geometric\\ Brownian Motion}
For each user $u$, the mean $\mu_u$ and variance $\sigma_u$ of her GBM trajectory are estimated from the sample mean and variance. 

Consider the set of all reviews $\langle d_t \rangle$ written by $u$, and $\langle e_t \rangle$ be the corresponding experience values of the reviews. 

Let $\widehat{m}_u=\frac{\sum_{d_t} log(e_t)}{|d_t|}$, and $\widehat{s}_u^2 = \frac{\sum_{d_t} (log(e_t)-\widehat{m}_u)^2}{|d_t-1|}$. 

Furthermore, let $\Delta$ be the average length of the time intervals for the reviews of user $u$. 

Now, $log(e_t) \sim N \big((\mu_u-\frac{\sigma_u^2}{2})\Delta +log(s_{0,u}), \sigma_u \sqrt{\Delta}\big)$.

From the above equations we can obtain the following estimates using Maximum Likelihood Estimation (MLE):
\begin{equation}
 \begin{aligned}
 \widehat{\sigma}_u &= \frac{\widehat{s}_u}{\sqrt{\Delta}}\\ 
 \widehat{\mu}_u &= \frac{\widehat{m}_u - log(s_{0,u})}{\Delta} + \frac{{\widehat{\sigma}_u}^2}{2}\\
  &= \frac{\widehat{m}_u - log(s_{0,u})}{\Delta} + \frac{{\widehat{s}_u}^2}{2\Delta} 
 \end{aligned}
\end{equation}

\subsubsection{Overall Processing Scheme}
Exploiting the results from the above discussions, the overall inference is an iterative process consisting of the following steps:
\begin{enumerate}\setlength{\itemsep}{0pt}
 \item Estimate facets $Z$ using Equation~\ref{eq:gibbs}.
 \item Estimate $\beta$ using Equations~\ref{eq:kalman-predict} and~\ref{eq:kalman-update}.
 \item Sort all reviews by timestamps, and estimate $E$ using Equation~\ref{eq:prop} and the Metropolis Hastings algorithm, for a random subset of the reviews.
 \item Once the experience values of all reviews have been determined, estimate $L$ using Equation~\ref{eq:word-exp}.
\end{enumerate}

\section{Experiments}
\label{sec:experiments}

We perform experiments with data from five communities in different domains:
\squishlist
 \item BeerAdvocate ({\tt \small beeradvocate.com}) and RateBeer\\({\tt \small ratebeer.com}) for beer reviews
 \item Amazon ({\tt\small amazon.com}) for movie reviews
 \item Yelp ({\tt \small yelp.com}) for food and restaurant reviews
 \item NewsTrust ({\tt \small newstrust.net}) for reviews of news media
\squishend

Table~\ref{tab:statistics} gives the dataset statistics\footnote{\small http://snap.stanford.edu/data/, http://www.yelp.com/dataset\_challenge/, http://resources.mpi-inf.mpg.de/impact/credibilityanalysis/data.tar.gz}. 
We have a total of $12.7$ million reviews from $0.9$ million users over $16$ years from all of the five communities combined. The first four communities are used for product reviews, from where we extract the following quintuple for our model $<userId, itemId, timestamp, rating, review>$. 
NewsTrust is a special community, which we discuss in Section~\ref{sec:usecases}.

\begin{table}
 \setlength{\tabcolsep}{1.4mm}
\begin{tabular}{lrrrc}
\toprule
\bf{Dataset} & \bf{\#Users} & \bf{\#Items} & \bf{\#Ratings} & \bf{\#Years}\\
\midrule
\bf{Beer (BeerAdvocate)} & 33,387 & 66,051 & 1,586,259 & 16\\
\bf{Beer (RateBeer)} & 40,213 & 110,419 & 2,924,127 & 13\\
\bf{Movies (Amazon)} & 759,899 & 267,320 & 7,911,684 & 16\\
\bf{Food (Yelp)} & 45,981 & 11,537 & 229,907 & 11\\
\bf{Media (NewsTrust)} & 6,180 & 62,108 & 89,167 & 9\\
\midrule
\bf{TOTAL} & 885,660 & 517,435 & 12,741,144 & -\\
\bottomrule
\end{tabular}
\vspace{-0.5em}
\caption{Dataset statistics.}
\label{tab:statistics}
\vspace{-2em}
\end{table}

\subsection{Data Likelihood, Smoothness\\ and Convergence}

Inference of our model is quite involved with different Markov Chain Monte Carlo methods. It is imperative to show that the resultant model is not only stable, but also improves the log-likelihood of the data. Although there are several measures to evaluate the quality of facet models, we report the following from~\cite{wallach}:\\
$LL = \sum_d \sum_{j=1}^{N_d} log\ P(w_{d,j} | \beta; \alpha)$. A higher likelihood indicates a better model.

Figure~\ref{fig:log-likelihood} contrasts the log-likelihood of the data from the continuous experience model and its discrete counterpart \cite{Subho:ICDM2015}. We find that the continuous model is stable and has a {\em smooth} increase in the data log-likelihood {\em per iteration}. This can be attributed to how smoothly the language model evolves over time, preserving the Markov property of experience evolution. Empirically our model also shows a fast convergence, as indicated by the number of iterations. 

On the other hand, the discrete model not only has a worse fit, but is also less smooth. It exhibits abrupt state transitions in the Hidden Markov Model, when the experience level changes (refer to Figure~\ref{fig:exp}). This leads to abrupt changes in the language model, as it is coupled to experience evolution. 


\subsection{Experience-aware Item Rating Prediction}

In the first task, we show the effectiveness of our model for item rating prediction. Given a user $u$, an item $i$, time $t$, and review $d$ with words $\langle w \rangle$ --- the objective is to predict the rating the user would assign to the item based on her {\em experience}. 

For prediction, we use the following features: The experience value $e$ of the user is taken as the last experience attained by the user during training. 
Based on the learned language model $\beta$, we construct the language feature vector $\langle F_w = log(max_z (\beta_{t,z,w})) \rangle$ of dimension $V$ (size of the vocabulary). That is, for each word $w$ in the review, we consider the value of $\beta$ corresponding to the best facet $z$ that can be assigned to the word at the time $t$. We take the log-transformation of $\beta$ which empirically gives better results.

Furthermore, as also done in the baseline works \cite{mcauleyWWW2013,Subho:ICDM2015}, we consider: $\gamma_g$, the average rating in the community;  $\gamma_u$, the offset of the average rating given by user $u$ from the global average; and $\gamma_i$, the rating bias for item $i$. 

Thus, combining all of the above, we construct the feature vector
\noindent $\langle \langle F_w \rangle, e, \gamma_g, \gamma_u, \gamma_i \rangle$ for each review with the user-assigned ground rating for training. We use Support Vector Regression~\cite{drucker97}, with the same set of default parameters as used in our discrete model~\cite{Subho:ICDM2015}, for rating prediction. 

\subsubsection{Baselines}

\begin{table*}[ht]
\centering
\begin{tabular}{lccccc}
\toprule
\bf{Models} & \bf{BeerAdvocate} & \bf{RateBeer} & \bf{NewsTrust} & \bf{Amazon} & \bf{Yelp}\\
\midrule
{\bf Continuous experience model} (this work) & {\bf 0.247} & {\bf 0.266} & {0.494} & {\bf 1.042} & {\bf 0.940}\\
Discrete experience model \cite{Subho:ICDM2015} & 0.363 & 0.309 & 0.{\bf 464} & 1.174 & 1.469\\
User at learned rate \cite{mcauleyWWW2013} & 0.379 & 0.336 & 0.575 & 1.293 & 1.732\\
Community at learned rate \cite{mcauleyWWW2013} & 0.383 & 0.334 & 0.656 & 1.203 & 1.534\\
Community at uniform rate \cite{mcauleyWWW2013} & 0.391 & 0.347 & 0.767 & 1.203 & 1.526\\
User at uniform rate \cite{mcauleyWWW2013} & 0.394 & 0.349 & 0.744 & 1.206 & 1.613\\
Latent factor model \cite{koren2011advances} & 0.409 & 0.377 & 0.847 & 1.248 & 1.560\\
\bottomrule
\end{tabular}
\vspace{-0.5em}
\caption{Mean squared error (MSE) for rating prediction. Our model performs better than competing methods.}
\label{fig:MSE}
\vspace{-1em}
\end{table*}

\begin{figure*}
\centering
 \includegraphics[width=\linewidth, height=4cm]{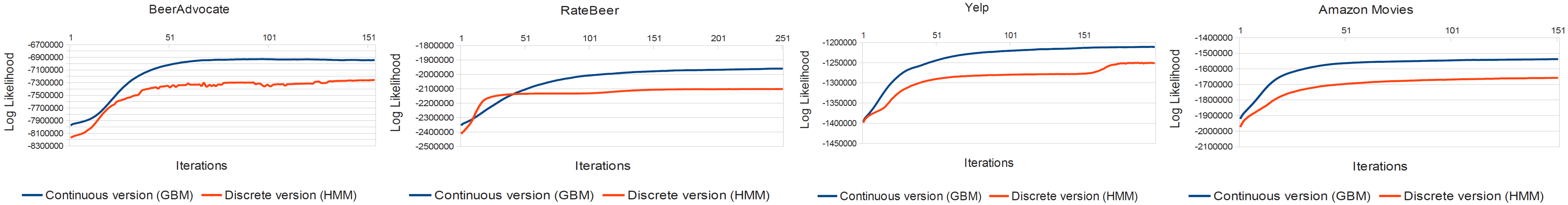}
    \vspace{-2em}
 \caption{Log-likelihood per iteration of discrete \cite{Subho:ICDM2015} vs. continuous experience model (this work).}
  \label{fig:log-likelihood}
  \vspace{-2em}
\end{figure*}

\begin{figure*}
\centering
 \includegraphics[width=\linewidth, height=4cm]{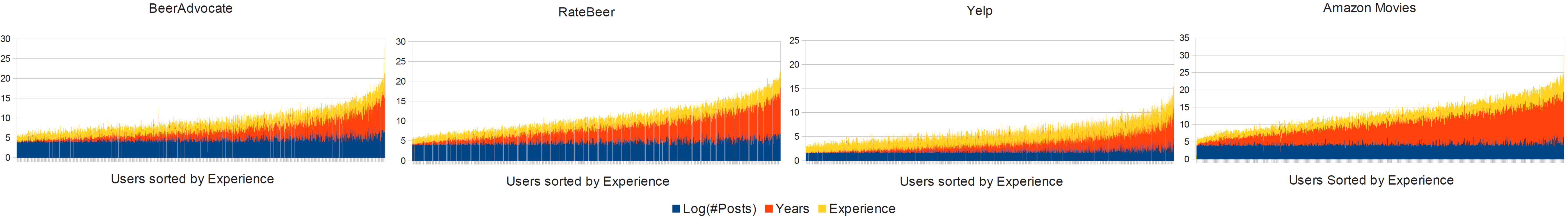}
    \vspace{-2em}
 \caption{Variation of {\em experience} ($e$) with {\em years} and {\em reviews} of each user. Each bar in the above stacked chart corresponds to a user with her most recent experience, number of years spent, and number of reviews posted in the community.}
  \label{fig:user-activity}
    \vspace{-2em}
\end{figure*}

\begin{figure*}
\centering
 \includegraphics[width=\linewidth, height=4cm]{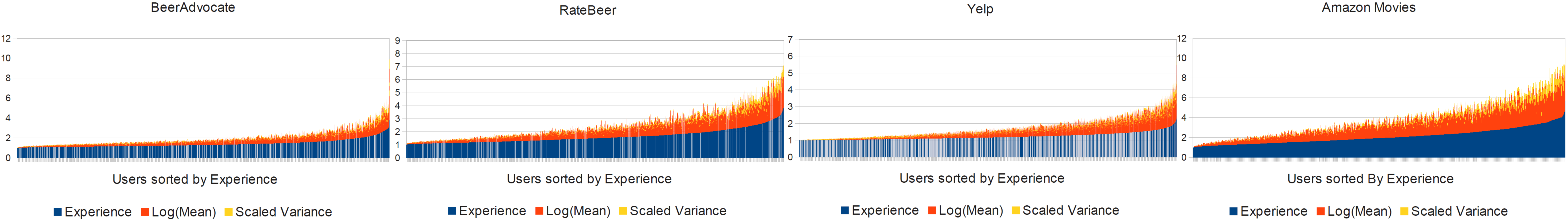}
    \vspace{-2em}
 \caption{Variation of {\em experience} ($e$) with {\em mean} ($\mu_u$) and {\em variance} ($\sigma_u$) of the GBM trajectory of each user ($u$). Each bar in the above stacked chart corresponds to a user with her most recent experience, mean and variance of her experience evolution.}
  \label{fig:user-mean}
    \vspace{-2em}
\end{figure*}

\begin{figure*}
\centering
 \includegraphics[width=\linewidth, height=4cm]{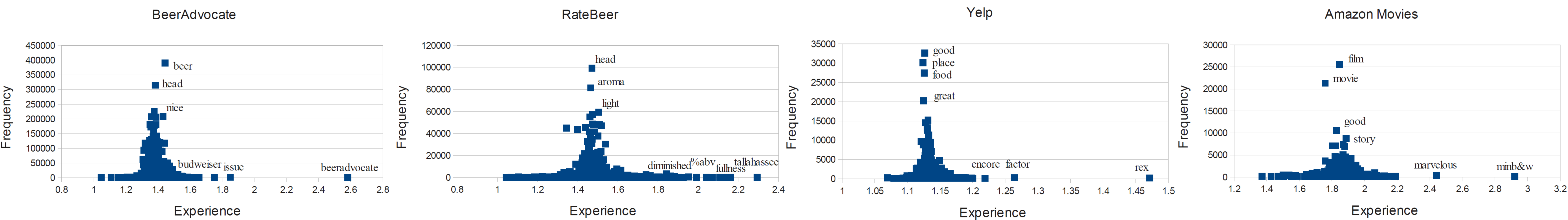}
    \vspace{-2em}
 \caption{Variation of {\em word frequency} with {\em word experience}. Each point in the above scatter plot corresponds to a word ($w$) in ``2011'' with corresponding frequency and experience value ($l_{t\text{=2011},w}$).}
  \label{fig:word-exp-freq}
    \vspace{-2em}
\end{figure*}

\begin{figure*}
\centering
 \includegraphics[width=\linewidth, height=4cm]{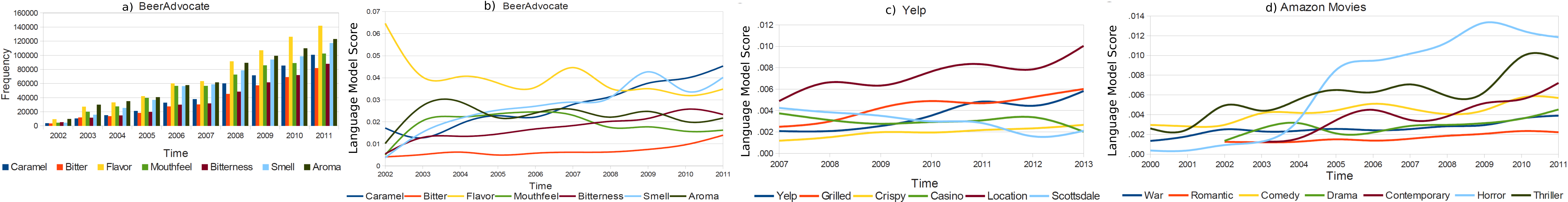}
    \vspace{-2em}
 \caption{{\em Language model} score ($\beta_{t,z,w} \boldsymbol{\cdot} l_{t,w}$) variation for sample words with {\em time}. Figure a) shows the count of some sample words over time in BeerAdvocate community, whose evolution is traced in Figure b). Figures c) and d) show the evolution in Yelp and Amazon Movies.}
  \label{fig:word-evol}
    \vspace{-2em}
\end{figure*}

We consider baselines [b -- e] from~\cite{mcauleyWWW2013}, and use their code\footnote{\small Code available from http://cseweb.ucsd.edu/~jmcauley/code/} for experiments. Baseline (f) is our prior discrete experience model\cite{Subho:ICDM2015}.

\squishlist
\item[a)]\emph{LFM}: A standard latent factor recommendation model~\cite{korenKDD2008}.
\item[b)]\emph{Community at uniform rate}: Users and products in a community evolve using a single ``global clock''~\cite{KorenKDD2010, xiongSDM2010, XiangKDD2010}, where the different stages of the community evolution appear at uniform time intervals. 
\item[c)]\emph{Community at learned rate}: This extends b) by learning the rate at which the community evolves with time, eliminating the uniform rate assumption.
\item[d)]\emph{User at uniform rate}: This extends b) to consider individual users, by modeling the different stages of a user's progression based on preferences and experience levels evolving over time. 
The model assumes a uniform rate for experience progression. 
\item[e)]\emph{User at learned rate}: This extends d) by allowing the {\em experience} of each user to evolve on a ``personal clock'', where the time to reach certain ({\em discrete}) experience levels depends on the user \cite{mcauleyWWW2013}. This is reportedly the best version of their experience evolution models.
\item[f)]\emph{Discrete experience model}: This is our prior work \cite{Subho:ICDM2015} on the discrete version of the experience-aware language model, where the experience of a user depends on the evolution of the user's maturing rate, facet preferences, and writing
style.
\squishend


\subsubsection{Quantitative Results}

Table~\ref{fig:MSE} compares the \emph{mean squared error (MSE)} for rating predictions in this task, generated by our model versus the six baselines. Our model outperforms all baselines --- except in the NewsTrust community, performing slightly worse than our prior work \cite{Subho:ICDM2015} (discussed in Section~\ref{sec:usecases}) --- reducing the MSE by ca. $11\%$ to $36\%$. Our improvements over the baselines are statistically significant at $99\%$ level of confidence determined by {\em paired sample t-test}.

For all models, we used the three most recent reviews of each user as withheld test data. All experience-based models consider the \emph{last} experience value reached by each user during training, and the corresponding learned parameters for rating prediction. Similar to the setting in~\cite{mcauleyWWW2013}, we consider users with a minimum of $50$ reviews. Users with less than $50$ reviews are grouped into a background model, and treated as a single user. We set $Z=5$ for BeerAdvocate, RateBeer and Yelp facets; and $Z=20$ for Amazon movies and $Z=100$ for NewsTrust which have richer latent dimensions. All {\em discrete} experience models consider $E=5$ experience levels. In the continuous model, the experience value $e \in (0, \infty)$. We initialize the parameters for our joint model as: $s_{0,u} = 1, \alpha=50/Z, \gamma=0.01$. Our performance improvement is strong for the {\em BeerAdvocate} community due to large number of reviews per-user for a long period of time, and low for NewsTrust for the converse.

\subsection{Qualitative Results}

{\noindent \bf User experience progression:} Figure~\ref{fig:user-activity} shows the variation of the users' {\em most recent} experience (as learned by our model), along with the number of reviews posted, and the number of years spent in the community. As we would expect, a user's experience increases with the amount of {\em time} spent in the community. On the contrary, number of reviews posted does not have a strong influence on experience progression. Thus, if a user writes a large number of reviews in a short span of time, her experience does not increase much; in contrast to if the reviews are written over a long period of time. 

Figure~\ref{fig:user-mean} shows the variation of the users' {\em most recent} experience, along with the mean $\mu_u$ and variance $\sigma_u$ of her Geometric Brownian Motion (GBM) trajectory --- all learned during inference. We observe that users who reach a high level of experience progress faster (i.e. a higher value of $\mu_u$) than those who do not. Experienced users also exhibit comparatively higher variance than amateur ones. This result also follows from using the GBM process, where the mean and variance tend to increase with time.

{\noindent \bf Language model evolution:} Figure~\ref{fig:word-exp-freq} shows the variation of the frequency of a word --- used in the community in ``2011'' --- with the {\em learned} experience value $l_{t,w}$ associated to each word. The plots depict a bell curve. Intuitively, the experience value of a word does not increase with general usage; but increases if it has been used by experienced users. Highlighted words in the plot give some interesting insights. For instance, the words ``beer, head, place, food, movie, story'' etc. are used with high frequency in the beer, food or movie community, but have an average experience value. On the other hand specialized words like ``beeradvocate, budweiser, \%abv, fullness, encore, minb\&w'' etc. have high experience value. 

Table~\ref{tab:sample-words} shows some top words used by {\em experienced} users and amateur ones in different communities, as learned by our model. Note that this is a ranked list of words with numeric values (not shown in the table). 
We see that experienced users are more interested about fine-grained facets like the mouthfeel, ``fruity'' flavors, and texture of food and drinks; narrative style of movies, as opposed to popular entertainment themes; discussing government policies and regulations in news reviews etc.

The word ``rex'' in Figure~\ref{fig:word-exp-freq} in Yelp, appearing with low frequency and high experience, corresponds to a user ``Rex M.'' with ``Elite'' status who writes humorous reviews with {\em self} reference.

Figure~\ref{fig:word-evol} shows the evolution of some sample words over {\em time} and experience (as given by our model) in different communities. The score in the {\em y-axis} combines the language model probability $\beta_{t,z,w}$ with the experience value $l_{t,w}$ associated to each word $w$ at time $t$. Figure~\ref{fig:word-evol} a) illustrates the frequency of the words in BeerAdvocate, while their evolution is traced in Figure~\ref{fig:word-evol} b). It can be seen that the overall usage of each word increases over time; but the evolution path is different for each word. For instance, the ``smell'' convention started when ``aroma'' was dominant; but the latter was less used by {\em experienced} users over time, and slowly replaced by (increasing use of) ``smell''. This was also reported in~\cite{DanescuWWW2013} in a different context. Similarly ``caramel'' is likely to be used more by {\em experienced} users, than ``flavor''. Also, contrast the evolution of ``bitterness'', which is used more by 
experienced users, compared to ``bitter''. 

In Yelp, we see certain food trends like ``grilled'' and ``crispy'' increasing over time; in contrast to a decreasing feature like ``casino'' for restaurants. For Amazon movies, we find certain genres like ``horror, thriller'' and ``contemporary'' completely dominating other genres in recent times.

\begin{table}
\scriptsize
 \setlength{\tabcolsep}{1.3mm}
\begin{tabular}{p{4.5cm}p{3.5cm}}
\toprule
{\bf Most Experience} & {\bf Least Experience}\\\midrule
{\noindent \bf BeerAdvocate} & \\
chestnut\_hued near\_viscous rampant\_perhaps faux\_foreign cherry\_wood sweet\_burning bright\_crystal faint\_vanilla boned\_dryness woody\_herbal citrus\_hops mouthfeel& originally flavor color didnt favorite dominated cheers tasted review doesnt drank version poured pleasant bad bitter sweet\\ \midrule
{\noindent \bf Amazon} & \\
aficionados minimalist underwritten theatrically unbridled seamless retrospect overdramatic diabolical recreated notwithstanding oblivious featurettes precocious & viewer entertainment battle actress tells emotional supporting evil nice strong sex style fine hero romantic direction superb living story\\\midrule
{\bf Yelp} &\\
rex foie smoked marinated savory signature contemporary selections bacchanal delicate grits gourmet texture exotic balsamic & mexican chicken salad love better eat atmosphere sandwich local dont spot day friendly order sit \\\midrule
{\bf NewsTrust} &\\
health actions cuts medicare oil climate major jobs house vote congressional spending unemployment citizens events & bad god religion iraq responsibility questions clear jon led meaningful lives california powerful\\
\bottomrule
\end{tabular}
\vspace{-1em}
\caption{Top words used by experienced and amateur users.}
\label{tab:sample-words}
\vspace{-2em}
\end{table}

\vspace*{-1mm}\section{Use-Case Study}
\label{sec:usecases}

So far we have focused on traditional item recommendation for items like beers or movies. Now we switch to different kind of items - newspapers and news articles - by analyzing the NewsTrust online community ({\tt \small newstrust.net}) (data available from~\cite{SubhoCIKM2015}). It features news stories posted and reviewed by members, 
some of whom are professional journalists and content experts. Stories are reviewed based on their objectivity, rationality, and general quality of language to present an unbiased and balanced narrative of an event with focus on \emph{quality journalism}. Unlike the other datasets, NewsTrust contains expertise of members that can be used as ground-truth for evaluating our model-generated {\em experience} values.

In our framework, each story is an item, which is rated and reviewed by a user. The facets are the underlying topic distribution of the reviews, with topics being {\em Healthcare, Obama Administration, NSA}, etc. The facet preferences can be mapped to the (political) polarity of users in this news community. 

\noindent {\bf Recommending News Articles.}
Our first objective is to recommend news to readers catering to their viewpoints, and experience. We apply our model with the same setting as with earlier datasets.
The mean squared error (MSE) results were reported in Section~\ref{sec:experiments}. Our model clearly outperforms most of the baselines; it performs only slightly worse regarding our prior work~\cite{Subho:ICDM2015} in this task --- possibly due to high rating sparsity in face of a large number of model parameters. 




\noindent {\bf Identifying Experienced Users.}
Our second task is to find experienced members of this community, who have the potential of being \emph{citizen journalists}. In order to evaluate the quality of the ranked list of experienced users generated by our model, we consider the following proxy measure for user experience. In NewsTrust, users have {\em Member Levels} determined by the NewsTrust staff based on  community engagement, time in the community, other users' feedback on reviews, profile transparency, and manual validation.
We use these member levels to categorize users as {\em experienced} or {\em inexperienced}. This is treated as the ground truth for assessing the ranking quality of our model against the baseline models~\cite{mcauleyWWW2013, Subho:ICDM2015} --- considering top $100$ users from each model ranked by experience. Here we consider the top-performing baseline models from the previous task.
We report the \emph{Normalized Discounted Cumulative Gain (NDCG)} 
and the {\em Normalized Kendall Tau Distance} 
for the ranked lists of users generated by all the models. The better model should exhibit higher {\em NDCG}, and lower {\em Kendall Tau Distance}. 
As Table~\ref{tab:ndcgUsers} shows, our model outperforms baseline models in capturing user maturity.

\begin{table}
\begin{center}
\small
 \setlength{\tabcolsep}{1.5mm}
 \begin{tabular}{p{3.8cm}p{1.1cm}p{2.6cm}}
 \toprule
\bf{Models} &\bf{NDCG} & \bf{Kendall Tau}\\
\bf{} & \bf{} & \bf{Normalized Distance}\\
\midrule 
{\bf Continuous experience model}\newline (this work) & {\bf 0.917} & {\bf 0.113} \\
Discrete experience model~\cite{Subho:ICDM2015} & 0.898 & 0.134 \\
User at learned rate~\cite{mcauleyWWW2013} & 0.872 & 0.180 \\
\bottomrule
 \end{tabular}
 \vspace{-1em}
\caption{Performance on identifying experienced users.}
\label{tab:ndcgUsers}
\end{center}
\vspace{-2em}
\end{table}

\section{Related Work}

State-of-the-art recommender systems \cite{korenKDD2008, koren2011advances} 
harness user-user and item-item similarities by means of latent factor models. 
Time-dependent phenomena such as bursts in item popularity, bias in ratings, 
and the temporal evolution of user community are investigated in~\cite{KorenKDD2010, xiongSDM2010, XiangKDD2010}. There is also prior work on anomaly detection~\cite{GunnemannGF14,Gunnemann2014}, capturing changes of social links~\cite{MaWSDM2011} and linguistic norms~\cite{DanescuWWW2013}. 
None of these prior works take into account the evolving experience and behavior of individual users.

Prior work that analyzed user review texts focused on sentiment analysis~\cite{linCIKM2009}, learning latent aspects and their ratings~\cite{wang2011, mukherjeeSDM2014, mcauleyRecSys2013}, and user-user interactions~\cite{West-etal:2014}. 
However, all of these prior approaches operate in a static, snapshot-oriented manner, without considering time at all.

The work~\cite{mcauleyWWW2013}, one of our baselines, has modeled and studied the influence of evolving user experience on rating behavior and for targeted recommendations. However, it disregards the vocabulary in the users' reviews. 
Our own recent work~\cite{Subho:ICDM2015} addressed this very limitation,
by means of language models that are specific to the experience level of an
individual user, and by modeling transitions between experience levels with a Hidden
Markov Model.
However, both of these works are limited to {\em discrete} experience levels leading to abrupt changes in both experience and language model. To address the above, and other related drawbacks, the current paper introduces continuous-time models for the smooth evolution of both
user experience and language model.

Wang et al.~\cite{Wang2006} modeled topics over time. However, the topics
themselves were constant, and time was only
used to better discover them. Dynamic topic models have been introduced by Blei et al. in 
\cite{BleiDTM,BleiCTM}. This prior work developed generic models
based on Brownian Motion, and applied them
to news corpora. 
\cite{BleiCTM} argues that the continuous model
avoids making choices for discretization and is also
more tractable compared to fine-grained discretization.
Our language model is motivated by the latter. We substantially extend it to capture evolving user behavior and experience
in review communities using Geometric Brownian Motion.

%

\vspace*{-1mm}\section{Conclusion}
In this paper,
we have proposed an experience-aware language model that can trace the {\em continuous} evolution of user experience and language explicitly over {\em time}.
We combine principles of Geometric Brownian Motion, Brownian Motion, and Latent Dirichlet Allocation to model a smooth temporal progression of user experience, and language model over time.
This is the first work to develop a continuous and generalized version of user experience evolution.


Our experiments -- with data from domains like beer, movies, food, and news -- demonstrate that our model effectively exploits user experience for item recommendation that
substantially reduces the mean squared error for predicted ratings, compared to the state-of-the-art baselines~\cite{mcauleyWWW2013,Subho:ICDM2015}. 

We further demonstrate the utility of our model in a use-case study about identifying experienced members in the NewsTrust community, where these users would be top candidates for being citizen journalists. Another similar use-case for our model can be to detect experienced medical professionals in the health community.


\vspace*{-1mm}
\balance
\bibliographystyle{plain}
\bibliography{kdd14}
\end{document}